\documentclass{article} 
\usepackage{nips15submit_e,times}
\usepackage{hyperref}
\usepackage{url}

\usepackage[pdftex]{pstricks}
\usepackage{verbatim}

\usepackage[square,sort&compress,comma,numbers]{natbib}

\usepackage[T1]{fontenc}

\usepackage{amsmath}
\usepackage{amsfonts}
\usepackage{amssymb}
\usepackage{algorithm}
\usepackage{algorithmic}
\usepackage{color}
\usepackage[official]{eurosym}

\usepackage[toc,page]{appendix}

\usepackage{subcaption}

\usepackage{mathtools}

\usepackage{arydshln}

\floatname{algorithm}{Procedure}

\let\oldhat\hat

\renewcommand{\hat}[1]{\oldhat{\mathbf{#1}}}

\usepackage{arydshln}

\usepackage{tikz}
\usepackage{verbatim}
\usetikzlibrary{fit}
\usetikzlibrary{backgrounds}
\usetikzlibrary{positioning}

\title{How to Discount Deep Reinforcement Learning:\\ Towards New Dynamic Strategies}

\author{
Vincent François-Lavet\\
University of Liege\\
\texttt{v.francois@ulg.ac.be} \\
\And
Raphael Fonteneau \\
University of Liege \\
\texttt{raphael.fonteneau@ulg.ac.be} \\
\AND
Damien Ernst \\
University of Liege \\
\texttt{dernst@ulg.ac.be} \\
}

\nipsfinalcopy 

\begin{document}
\maketitle

\begin{abstract}
Using deep neural nets as function approximator for reinforcement learning tasks have recently been shown to be very powerful for solving problems approaching real-world complexity such as \cite{mnih2015human}. 
Using these results as a benchmark, we discuss the role that the discount factor may play in the quality of the learning process of a deep Q-network (DQN). When the discount factor progressively increases up to its final value, we empirically show that it is possible to significantly reduce the number of learning steps. When used in conjunction with a varying learning rate, we empirically show that it outperforms original DQN on several experiments. 
We relate this phenomenon with the instabilities of neural networks when they are used in an approximate Dynamic Programming setting. We also describe the possibility to fall within a local optimum during the learning process, thus connecting our discussion with the exploration/exploitation dilemma.
\end{abstract}

\section{Introduction}

Reinforcement learning is a learning paradigm aiming at learning optimal behaviors while interacting within an environment \cite{sutton1998introduction}. One of the main challenge met when designing reinforcement learning algorithms is the fact that the state space may be very large or continuous, potentially leading to the fact that state(-action) value functions may not be represented comprehensively (for instance, using lookup tables) \cite{Busoniu2010}. Recently, it has been empirically demonstrated in \cite{mnih2015human} that using deep neural networks as a function approximator may be very powerful in situations with high-dimensional sensory inputs such as the Atari games benchmark \cite{bellemare2012arcade}. However, one of the main drawback of using (deep) neural networks as function approximator is that they may potentially become unstable when combined with a Q-Learning-type recursion \cite{baird1995residual}.

In this paper, we propose to discuss the role that the discount factor may play in the stability and convergence of deep reinforcement learning algorithms. We empirically show that an increasing discount factor has the potential to improve the quality of the learning process. We also discuss the link between the discount factor influence, the instabilities of (deep) neural networks and the vanilla exploration/exploitation dilemma. 

A motivation for this work comes from empirical studies about the attentional and cognitive mechanisms in delay of gratification. One well known experiment in the domain was a series of studies in which a child was offered a choice between one small reward provided immediately or two small rewards if they waited for a short period ("marshmallow experiment" \cite{mischel1972cognitive}).  The capacity to wait longer for the preferred rewards seems to develop markedly only at about ages 3-4. By 5 years old, most children are able to demonstrate better self-control by gaining control over their immediate desires. According to this theory, it seems plausible that following immediate desires at a young age is a better way to develop its abilities and that delaying strategies is only advantageous afterwards when pursuing longer term goals. Similarly, reinforcement learning may also have an advantage of starting to learn by maximizing rewards on a short-term horizon and progressively giving more weights to delayed rewards.

The remaining of the paper is organized the following: we first recall the main equations used in the deep reinforcement learning problem formulation originally introduced in \cite{mnih2015human}. We then discuss the factors that influence the stability of (deep) neural networks. In light of this analysis, we suggest the possibility to modify the discount factor along the way to convergence up to its final value in order to speed up the learning process. 
To illustrate our approach, we use the benchmark proposed in \cite{mnih2015human}. In this context, we then discuss the role of the learning rate as well as the level of exploration. 



\section{Problem Formulation}

We consider tasks in which an agent interacts with an environment so as to maximize expected future rewards 
\begin{equation}
Q^*(s,a)=\operatorname*{max}_{\pi}\mathbb{E} [r_t, \gamma r_{t+1}, \gamma^2 r_{t+2}+ \ldots | s_t=s, a_t=a, \pi ]
\label{def_Q}
\end{equation}
which is the maximum sum of rewards $r$ discounted by $\gamma$ at each time-step $t$, achievable by a behaviour stochastic policy $\pi : S \times A \rightarrow [0,1]$ where $\pi(s,a)$ denotes the probability that action $a$ may be chosen by policy $\pi$ in state $s$. 
In this paper, the task is described by a time-invariant stochastic discrete-time system whose dynamics can be described by the following equation:
\begin{eqnarray}
s_{t+1} = f(s_{t},a_{t},w_{t})
\end{eqnarray}
where for all t, $s_t$ is an element of the state space $S$, the action $a_t$ is an element of the action
space $A$ and the random disturbance $w_t$ is an element of the disturbance space $W$ generated by a time-invariant conditional probability distribution $w_t \sim P(. | s, a)$. In our experiments, the dynamics will be given by the Atari emulator, the state space will be based on observed pixels of the screen 
and the action space is the set of legal game actions $A=\{1, . . . ,K\}$. 


The (unique) solution of the Bellman equation for the Q-value function \cite{bellman1962applied} is given by:
\begin{equation}
Q^*(s, a) = (H Q^*)(s, a) 
\label{Bellman_Q}
\end{equation}
where $H$ is an operator mapping any function $K : S \times A \rightarrow \mathbb{R}$ and defined as follows:
\begin{equation}
(HK)(s,a) = \mathbb{E}[r(s,a,w)+\gamma \operatorname*{max}_{a' \in A} K(f(s,a',w),a')]
\label{op_mapping_Bellman_Q}
\end{equation}

\begin{figure}[ht!]
 \centering
\begin{tikzpicture}
 [	NN/.style={circle,draw=blue!50,fill=blue!20,thick, inner sep=5pt,minimum size=6mm},
   	inputs/.style={rectangle,draw=black!50,fill=black!20,thick, inner sep=5pt,minimum size=4mm},
   	target/.style={rectangle,draw=black!50,fill=green!20,thick, inner sep=5pt,minimum size=4mm},
 	Vhat/.style={circle,draw=black!50,fill=black!20,thick, inner sep=5pt,minimum size=4mm},
 	ddt/.style={rectangle,draw=black!50,thick, inner sep=5pt,minimum size=4mm}];
 
  \node at (0,0) 	[NN, align=center] (NN) {Update \\$Q(s,a;\theta)$};
  \node [NN, right=of NN, align=center] (NN_t) {Every C:\\$\theta^{-}:=\theta$};
  \node [left=of NN,yshift=-7mm, xshift=-8mm,] 	[inputs] (rt) {$r_{1}, \ldots, r_{N_{replay}}$};
  \node [above=of rt, yshift=-7mm] 	[inputs] (st) {$s_{1}, \ldots, s_{N_{replay}}, a_{1}, \ldots, a_{N_{replay}}$};
  \node [below=of rt,yshift=7mm] 	[inputs] (st1) {$s_{1+1}, \ldots, s_{N_{replay+1}}$};
  \node [below=of NN,yshift=-2mm] 	[target] (target) {$r_t + \gamma \underset{a' \in A}{\operatorname{max}} ( Q(s_{t+1},a'; \theta^{-}) )$};
  \node[inner sep=0,minimum size=0,above=of NN, yshift=-3mm] (k) {}; 
  \node[draw,left=of rt, xshift=-8mm, align=center] (k3) {Policy}; 
  \node[inner sep=0,minimum size=0] (k1) at (k -| k3) {}; 
  \node[inner sep=0,minimum size=0] (k2) at (target -| k3) {}; 

 \draw [->] (st.east) -- (NN.west) node[midway, above] {$i^l$};
 \draw [->] (st1.east) -- (target);
 \draw [->] (rt.east) -- (target);
 \draw [->] (target) -- (NN) node[midway, right] {$o^l$};
 \draw [-] (NN) -- (k);
 \draw [-] (k) -- (k1);
  \draw [-] (k1) -- (k3);
  \draw [->] (k3) -- (st.west);  \draw [->] (k3) -- (rt.west);  \draw [->] (k3) -- (st1.west);
  \draw [->] (NN) -- (NN_t);
  \draw [->] (NN_t) -- (target);   

\end{tikzpicture}
 \caption{Sketch of the algorithm. $Q(s,a;\theta)$ is initialized to close to 0 everywhere on its domain; $N_{replay}$ is the size of the replay memory; the target Q-network parameters $\theta^-$ are only updated every C iterations with the Q-network parameters $\theta$ and are held fixed between updates; the variable $i^l$ corresponds to a mini-batch of tuples $(s, a)$ taken randomly in the replay memory and the variable $o^l$ is the corresponding target value for each tuple.}
 \label{ONFQ_schema}
\end{figure}
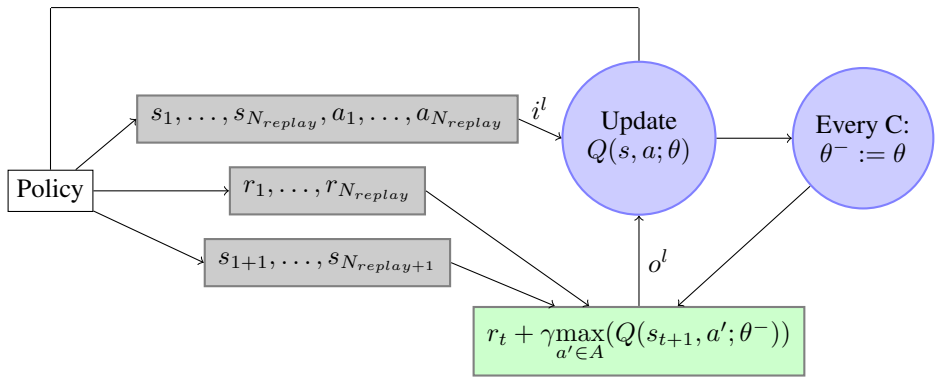

For most problems approaching real-world complexity, we need to learn a parameterized value function $Q(s, a; \theta_k)$. 
General function-approximation system such as neural networks are well suited to deal with high-dimensional sensory inputs. In addition, they work readily online, i.e. they can make use of additional samples obtained as learning happens by using only one supervised learning problem. 
In the case where a neural network $Q(s,a;\theta_k)$ is used to aim at convergence towards $Q^*(s,a)$, the parameters $\theta_k$ may be updated by stochastic gradient descent (or a variant), updating the current value $Q(s , a ; \theta_k )$ towards a target value $Y_k^Q = r(s,a,w) +\gamma \operatorname*{arg\,max}_{a' \in A} Q(f(s,a',w),a';\theta_k^{-})$ where $\theta_k^{-}$ refers to parameters from some previous Q-network. The Q-learning update when using the squared-loss amounts in updating the weights :
\begin{equation}
\theta_{k+1}=\theta_{k}+\alpha (Y_k^Q - Q (s,a; \theta_k)) \nabla_{\theta_k} Q(s ,a ; \theta_k)
\label{QlearningitNN}
\end{equation}
where $\alpha$ is a scalar step size called the learning rate. 
A sketch of the algorithm is given in Figure \ref{ONFQ_schema}.

\section{Instabilities of the online neural fitted Q-learning}


The Q-learning rule from Equation \ref{QlearningitNN} can be directly implemented online using a neural network (as is the case for the Deep Q-Network). However, due to the generalization and extrapolation abilities of neural nets, they can build unpredictable changes at different places in the state-action space. It is known that errors may be propagated and that this may even become unstable.  Additional care must therefore be taken since it can not be guaranteed that the current estimation for the accumulated costs always underestimates the optimal cost, and therefore convergence is not assured \cite{tsitsiklis1997analysis,gordon1999approximate}. These convergence problems are verified experimentally and it has been reported that convergence may be slow or even unreliable with this online update rule \cite{riedmiller2005neural}. 

We offer an analysis based on the complexity of the policy class \cite{jiang2015dependence} via a parallel with machine learning. High complexity machine learning techniques have the ability to represent their training set well, but  at risk of overfitting. In contrast, models with lower complexity do not tend to overfit, but may fail to capture important features. 
In the reinforcement learning setting, an equivalence with the model complexity of the learning method may be seen in the policy class complexity. Specifically, in the context of online neural fitted reinforcement learning, the discount factor as well as the neural network architecture control the number of possible policies. Similarly to the bias-variance tradeoff observed in supervised learning, reinforcement learning also faces a trade-off between large policy class complexity and low policy class complexity. 

It is already known that the optimal solution may actually be found in the set of policy with a planning horizon $\gamma$ smaller than the evaluation horizon $\gamma_{eval}$ specified by the problem formulation in case of only inaccurate information on the actual dynamics is available a priori (e.g. in the case where a small number of tuples is available) \cite{jiang2015dependence}.  It is also well known that the longer the planning horizon, the greater the computational expense of computing an optimal policy \cite{kearns2002sparse}. 

In the case of neural fitted value learning, the difficulty to target a high policy class complexity is severe because targeting a high discount factor leads to propagation of errors and instabilities. 
Some practical ways to prevent instabilities make use of a replay memory, clipping the error term, a separate target Q-network in Equation \ref{QlearningitNN} and a convolutional network architecture \cite{mnih2015human}. Using the double Q-learning algorithm also help reducing overestimations of Q-value function 
caused by the generalization exageration of the regression method \cite{dqn_doubleQlearning}.


In this paper, we investigate the tradeoff between performance of the targeted policy versus stability and speed of the learning process. 
We investigate the possibility to reduce instabilities during learning by working on an adaptive discount factor so as to soften the errors through learning. The goal is to target a high policy class complexity while reducing error propagations during the Deep Q learning iterations. 

\vspace{1cm}

\section{Experiments}

\begin{figure}[!htb]
\centering
  \includegraphics[width=200px]{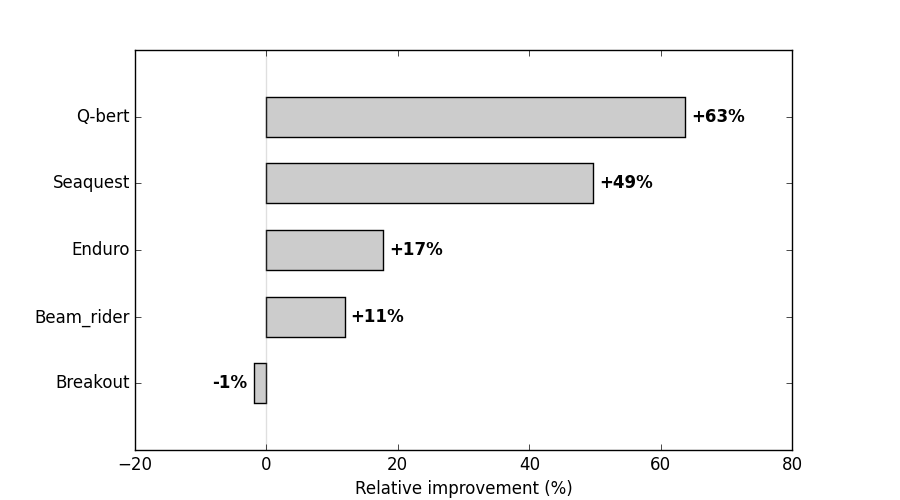}
  \caption{Summary of the results for an increasing discount factor. Reported scores are the relative improvements after 20M steps between an increasing discount factor and a constant discount factor set to its final value $\gamma=0.99$. 
 Details are given in Appendix - Table \ref{tab:results_inc_gamma}.}
  \label{fig:results_inc_gamma}
\end{figure}

The deep Q-learning algorithm described in \cite{mnih2015human} is used as a benchmark. All hyperparameters are kept identical if not stated differently. The main modification comes from the discount factor which is increased at every epoch (250 000 steps) with the following formula:

\begin{equation}
\gamma_{k+1}=1 - 0.98(1-\gamma_k)
\label{eq:inc_gamma_rule}
\end{equation}
The learned policies are then evaluated for 125000 steps with an $\epsilon$-greedy policy identical to the original benchmark with $\epsilon_{test}=0.05$. 
The reported scores are the highest average episode score of each simulation where these evaluation episodes were not truncated at 5 min. Each game is simulated 5 times for each configuration with varying seeds and results are reported in Figure \ref{fig:results_inc_gamma}. It can be observed that by simply using an increasing discount factor, learning is faster for four out of the five tested games and similar for the remaining game. We conclude that by starting with a low discount factor, we obtain faster policy improvement thanks to less instability. In addition, this also provides more robustness with respect to neural network weights initialisation. 


\subsection{Convergence of the neural network}
\begin{figure}[!htb]
\centering
\minipage{0.42\textwidth}
  \includegraphics[width=\linewidth]{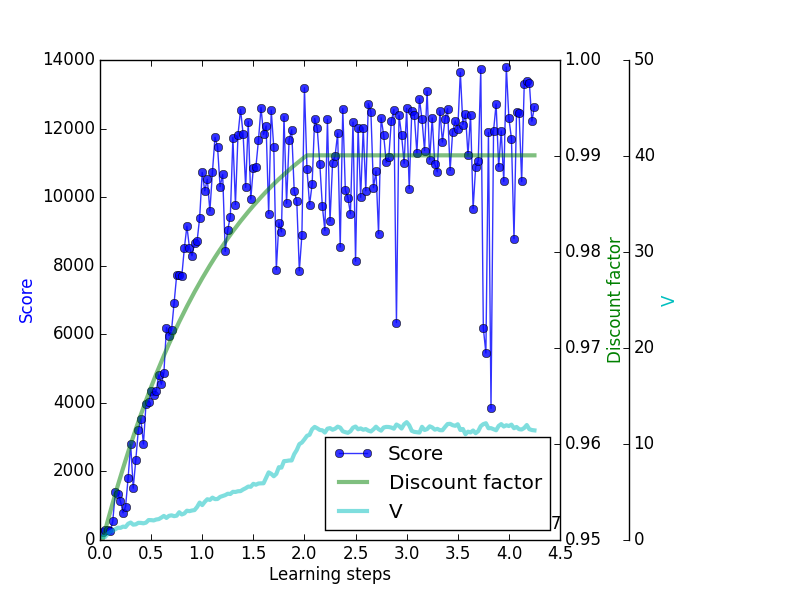}
\endminipage
\minipage{0.42\textwidth}
  \includegraphics[width=\linewidth]{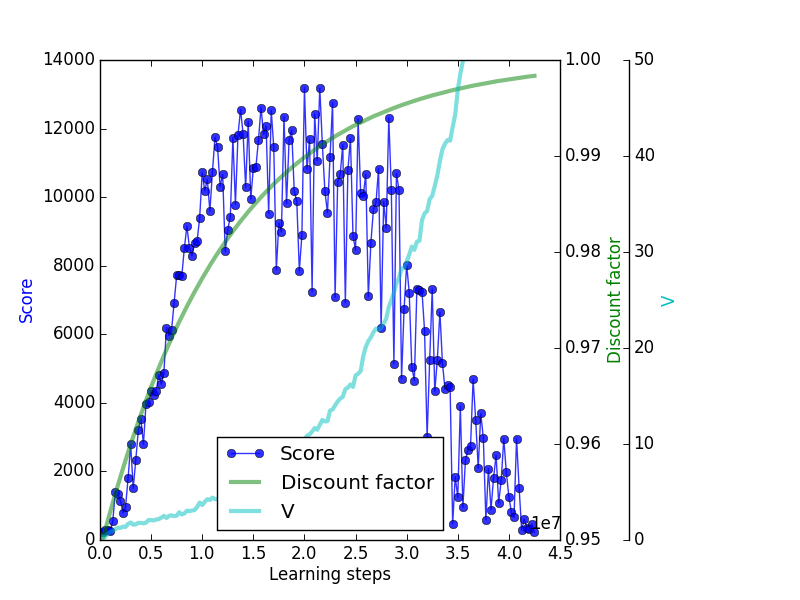}
\endminipage
\\
\minipage{0.42\textwidth}
  \includegraphics[width=\linewidth]{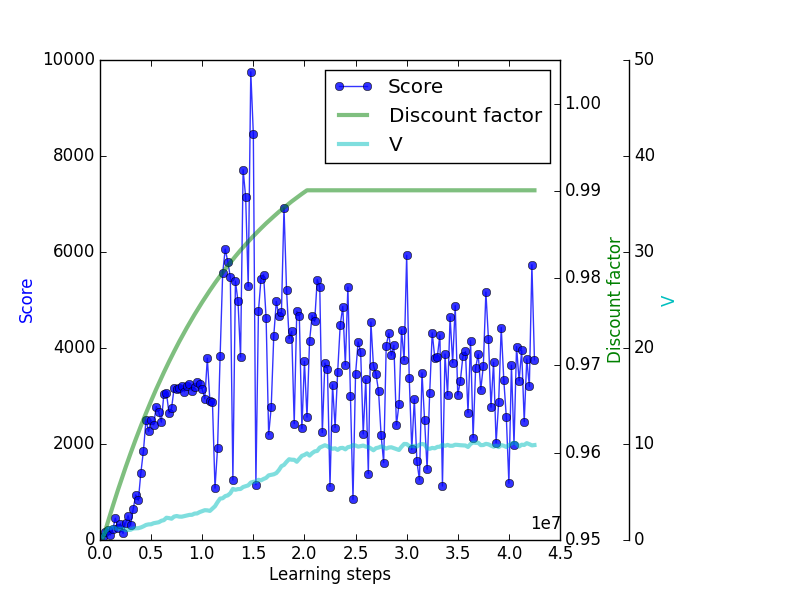}
\endminipage
\minipage{0.42\textwidth}
  \includegraphics[width=\linewidth]{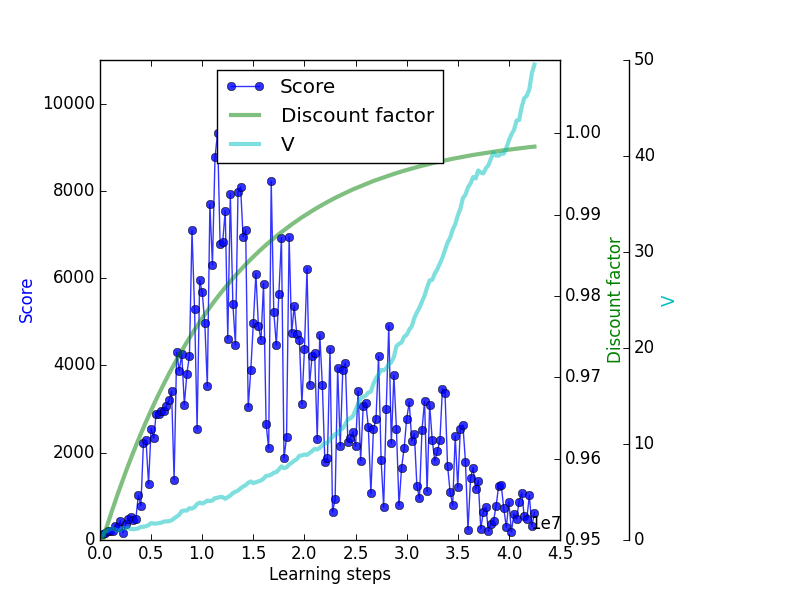}
\endminipage

  \caption{Illustration for the game q-bert (top) and seaquest (bottom) for a discount factor $\gamma$ kept at 0.99 after 20M learning steps (on the left) as well as a discount factor that keeps increasing to a value close to 1.}
  \label{ref:high_gamma}
\end{figure}

We now discuss the stability of DQN. We use the experimental rule from Equation \ref{eq:inc_gamma_rule} and either let $\gamma$ increase or keep it constant when it attains 0.99 (at 20M steps).
This is illustrated on Figure \ref{ref:high_gamma}.
It is shown that increasing $\gamma$ without additional care degrades severely the score obtained beyond $\gamma \approx 0.99$. 
By looking at the average V value, it can be seen that overestimation is particularly severe which causes the poor policies.


\subsection{Further improvement with an adaptive learning rate}
Since instabilities are only severe when the discount factor is high, we now study the possibility to use a more aggressive learning rate in the neural network when working at a low discount factor because potential errors would have less impact at this stage. The learning rate is then reduced along with the increasing discount factor so as to end up with a stable neural Q-learning function. 

\begin{figure}[!htb]
\centering
  \includegraphics[width=200px]{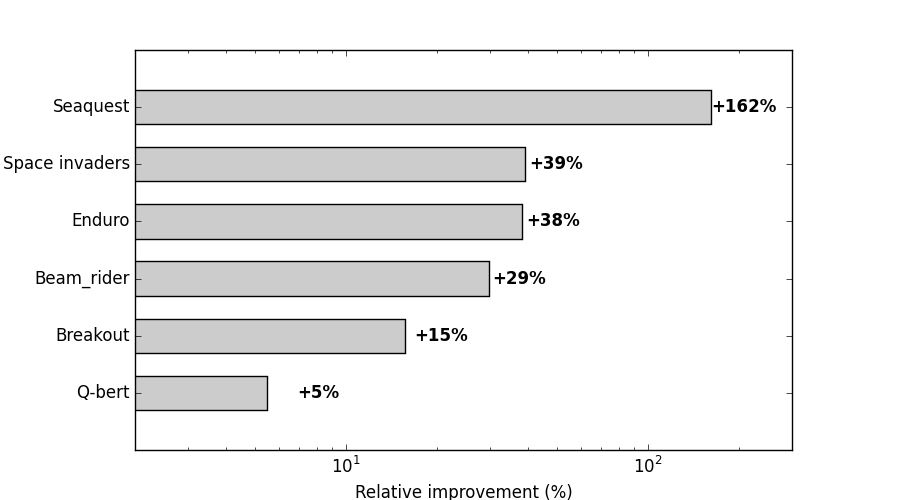}
  \caption{Summary of the results for a decreasing learning rate. Reported scores are the improvement after 50M steps when using both a dynamic discount factor and a dynamic learning rate. Details are given in  Appendix - Table \ref{tab:results_dec_lr}.}
  \label{fig:r}
\end{figure}

\begin{figure}[!htb]
\centering
\minipage{0.38\textwidth}
  \includegraphics[width=\linewidth]{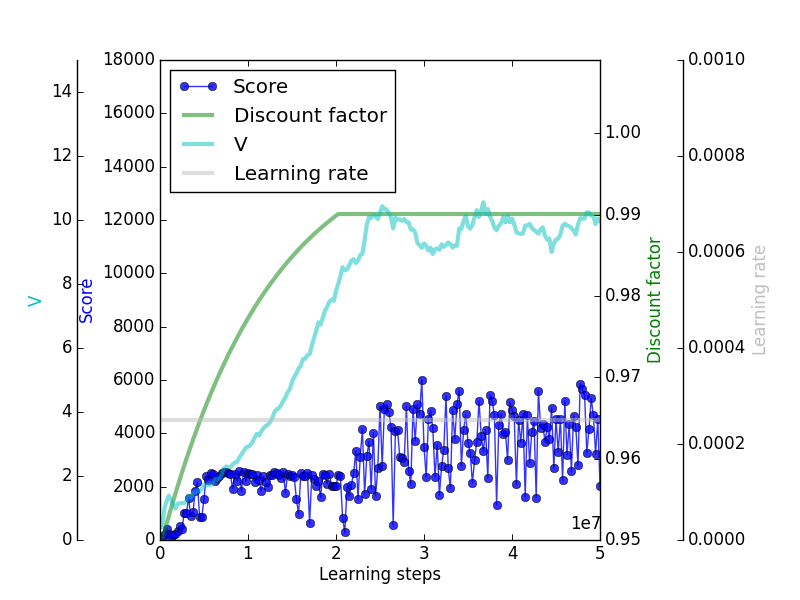}
\endminipage
\minipage{0.38\textwidth}
  \includegraphics[width=\linewidth]{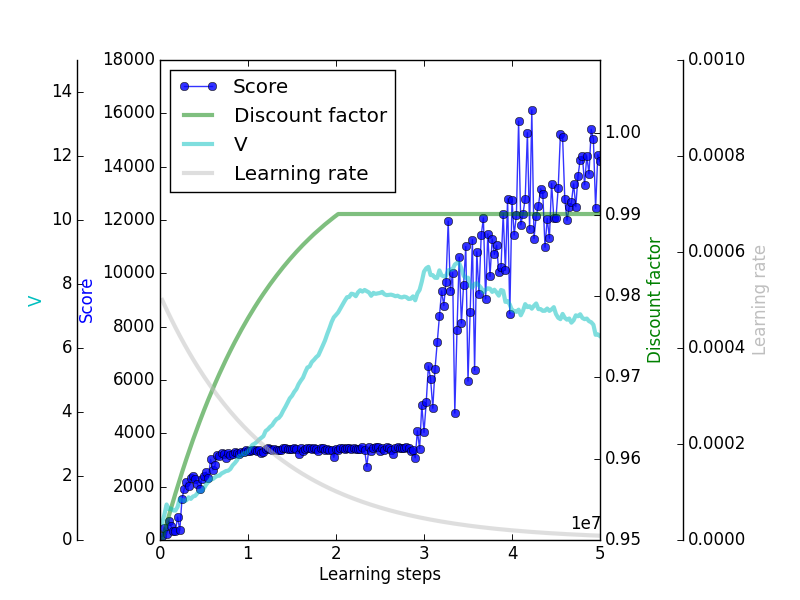}
\endminipage
\\
\minipage{0.38\textwidth}
  \includegraphics[width=\linewidth]{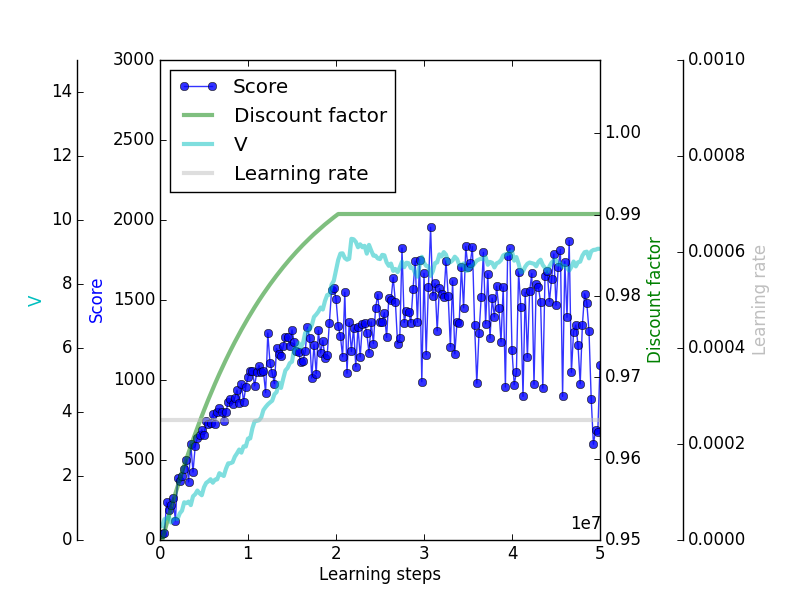}
\endminipage
\minipage{0.38\textwidth}
  \includegraphics[width=\linewidth]{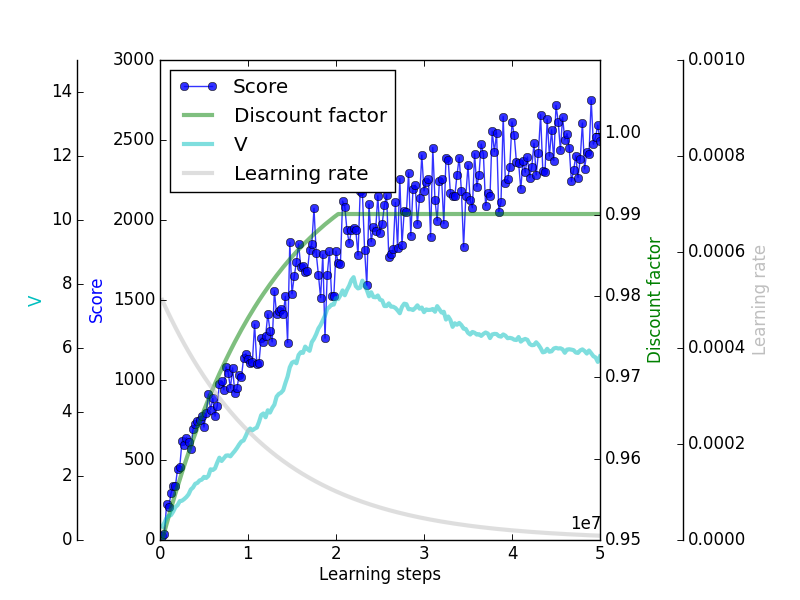}
\endminipage

  \caption{Illustration for the game seaquest (top) and space invaders (bottom). On the left, the deep Q-network with original parameters ($\alpha=0.00025$) and on the right with a decreasing learning rate.}
  \label{fig:dec_lr}
\end{figure}

We start with a learning rate of 0.005, i.e. twice as big as the one considered in the original benchmark and we use the following simple rule at every epoch:
$$\alpha_{k+1}=0.98 \alpha_{k} $$
With $\gamma$ that follows Equation \ref{eq:inc_gamma_rule} and is kept constant when it attains 0.99, we manage to improve further the score obtained on all of the games tested. Results are reported in Figure \ref{fig:r} and an illustration is given for two different games in Figure \ref{fig:dec_lr}. It can be noted that the value function V decreases when $\gamma$ is hold fixed and when the learning rate is lowered which is a sign of a decrease of the overestimations of the Q-value function.

\subsection{Exploration / Exploitation Dilemma}

Errors in the policy may also have positive impacts since it increases exploration \cite{kaelbling1996reinforcement}. When using a lower discount factor, it actually decreases exploration and opens up the risk of falling in a local optimum in the value iteration learning. This is observed as the agent gets repeatedly a score lower than the optimal while being unable to discover some parts of the state space. 

\begin{figure}[!htb]
\centering
\minipage{0.42\textwidth}
  \includegraphics[width=\linewidth]{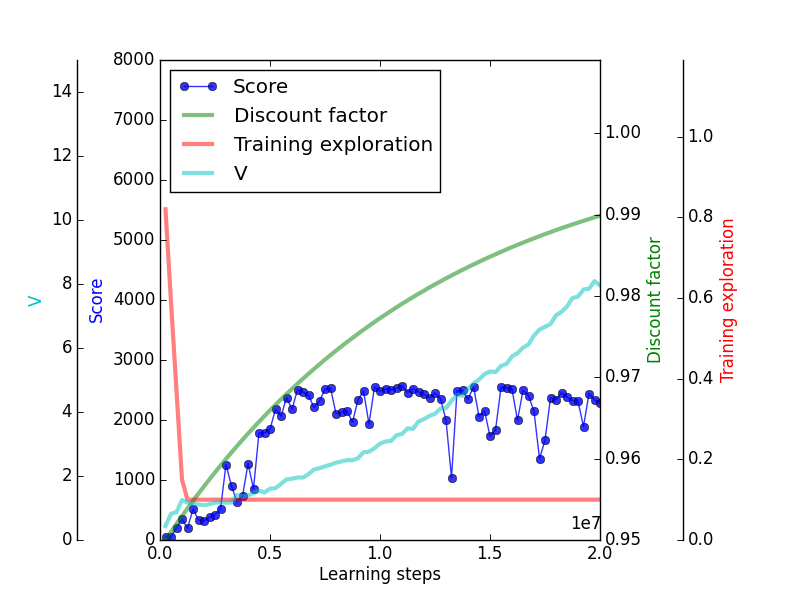}
\endminipage
\minipage{0.42\textwidth}
  \includegraphics[width=\linewidth]{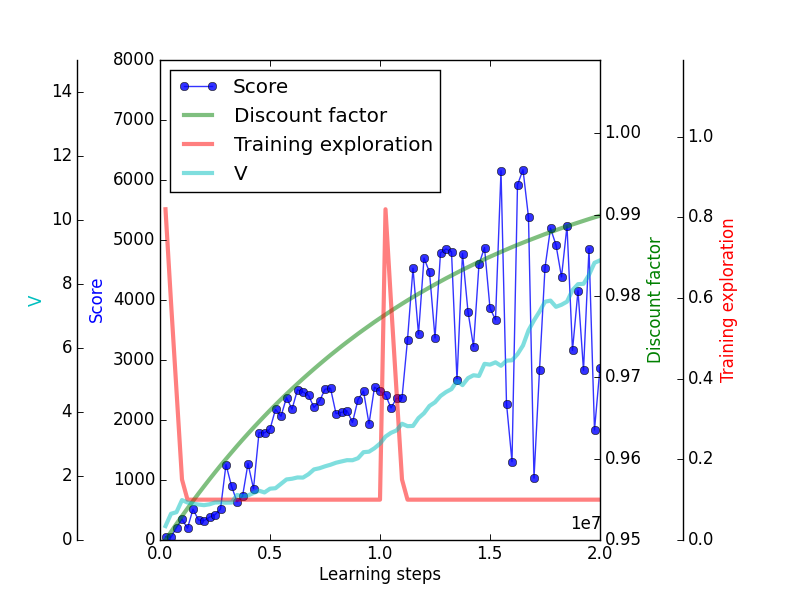}
\endminipage

  \caption{Illustration for the game seaquest for a case where the flat exploration rate fails to get out of a local optimum on the left. And on the right illustration where a simple rule that increases exploration allows to get out of the local optimum.}
  \label{seaquest_explo}
\end{figure}

In this case, an actor-critic-type algorithm that increases the level of exploration 
may allow to overcome this problem. 
We believe that an actor critic agent that also manages adaptively the level of exploration is important to further improve deep reinforcement learning algorithms. In order to illustrate this, a simple rule has been applied in the case of the game seaquest as can be seen on Figure \ref{seaquest_explo}. This rule adapts the exploration during the training process in the $\epsilon$-greedy action selection until the agent was able to get out of the local optimum. 

\subsection{Towards an actor-critic algorithm}
\begin{figure}[ht!]
 \centering
 
\begin{tikzpicture}
 [	NN/.style={circle,draw=blue!50,fill=blue!20,thick, inner sep=5pt,minimum size=6mm},
   	inputs/.style={rectangle,draw=black!50,fill=black!20,thick, inner sep=5pt,minimum size=4mm},
   	target/.style={rectangle,draw=black!50,fill=green!20,thick, inner sep=5pt,minimum size=4mm},
 	Vhat/.style={circle,draw=black!50,fill=black!20,thick, inner sep=5pt,minimum size=4mm},
 	ddt/.style={rectangle,draw=black!50,thick, inner sep=5pt,minimum size=4mm}];
 
  \node at (0,0) 	[NN, align=center] (NN) {Update \\$Q(s,a;\theta)$};
  \node [NN, right=of NN, align=center] (NN_t) {Every C:\\$\theta^{-}:=\theta$};
  \node [left=of NN,yshift=-7mm, xshift=-8mm,] 	[inputs] (rt) {$r_{1}, \ldots, r_{N_{replay}}$};
  \node [above=of rt, yshift=-7mm] 	[inputs] (st) {$s_{1}, \ldots, s_{N_{replay}}, a_{1}, \ldots, a_{N_{replay}}$};
  \node [below=of rt,yshift=7mm] 	[inputs] (st1) {$s_{1+1}, \ldots, s_{N_{replay+1}}$};
  \node [below=of NN,yshift=-2mm] 	[target] (target) {$r_t + \gamma \underset{a' \in A}{\operatorname{max}} ( Q(s_{t+1},a'; \theta^{-}) )$};
  \node[inner sep=0,minimum size=0,above=of NN, yshift=-3mm] (k) {}; 
  \node[draw,left=of rt, xshift=-8mm, align=center] (k3) {Policy\\ \& Update\\ of $\gamma$, $\alpha$ and $\epsilon$}; 
  \node[inner sep=0,minimum size=0] (k1) at (k -| k3) {}; 
  \node[inner sep=0,minimum size=0] (k2) at (target -| k3) {}; 

 \draw [->] (st.east) -- (NN.west) node[midway, above] {$i^l$};
 \draw [->] (st1.east) -- (target);
 \draw [->] (rt.east) -- (target);
 \draw [->] (target) -- (NN) node[midway, right] {$o^l$};
 \draw [-] (NN) -- (k);
 \draw [-] (k) -- (k1);
  \draw [-] (k1) -- (k3);
  \draw [-] (k3) -- (k2);
  \draw [->] (k3) -- (st.west);  \draw [->] (k3) -- (rt.west);  \draw [->] (k3) -- (st1.west);
  \draw [->] (k2) -- (target);   
  \draw [->] (NN) -- (NN_t);
  \draw [->] (NN_t) -- (target);   

\end{tikzpicture}
 \caption{Sketch of the actor-critic-type algorithm for deep reinforcement learning that manages adaptively the discount rate $\gamma$, the learning rate $\alpha$ as well as the level of exploration $\epsilon$.}
 \label{ONFQ_schema_general}
\end{figure}
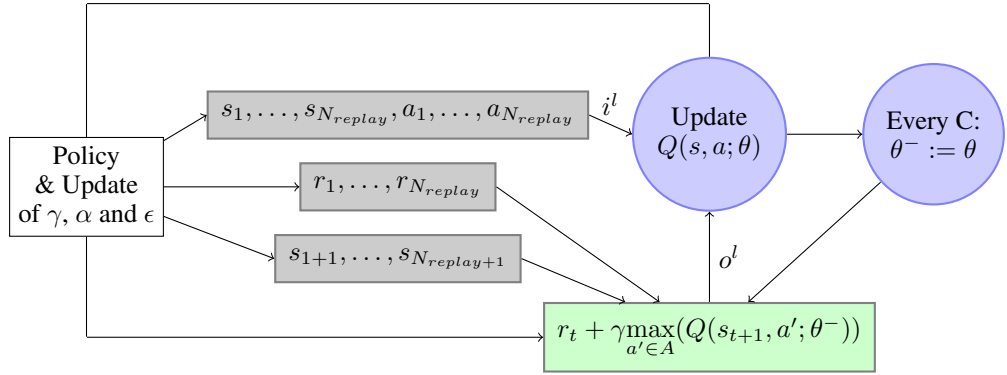

Following the ideas discussed above, Figure \ref{ONFQ_schema_general} represents the general update scheme that we propose to further improve the performance of deep reinforcement learning algorithms.

\section{Conclusion}

This paper introduced an approach to speed-up the convergence and improve the quality of the learned Q-function in deep reinforcement learning algorithms. It works by adapting the discount factor and the learning rate along the way to convergence. We used the deep Q-learning algorithms for Atari 2600 computer games as a benchmark and our approach showed improved performances for the 6 tested games. These results motivate further experiments, in particular it would be interesting to develop an automatic way to adapt online the discount factor along with the learning rate and possibly the level of exploration. It would also be of interest to combine this approach with the recent advances in deep reinforcement learning : the massively parallel architecture "Gorilla" \cite{nair2015massively}, the double Q-learning algorithm \cite{dqn_doubleQlearning}, the prioritized experience replay \cite{schaul2015prioritized}, the dueling network architecture \cite{wang2015dueling} or a recurrent architecture such as \cite{hausknecht2015deep}.



\appendix

\bibliographystyle{unsrt} 
\bibliography{references}{} 

\newpage
\appendix
\section{Appendix}

\begin{table}[h]
\begin{center}
\begin{tabular}{  | l  | c | c | c | c | c | }
  \hline                       
    & Seaquest & Q-bert & Breakout & Enduro & Beam rider \\
  \hline                       
  $\gamma=0.99$, 20M steps & 4766 & 7930 & 346 & 817 & 8379 \\
  & 4774 & 8262 & 359 & 950 & 9013 \\
  & 2941 & 7527 & 372 & 821 & 9201 \\
  & 3597 & 8214 & 374 & 863 & 8496 \\
  & 4280 & 7867 & 365 & 777 & 8408 \\
\hdashline
  \multicolumn{1}{|r|}{Average}  & \textbf{4072} & \textbf{7960} & \textbf{363} & \textbf{846} & \textbf{8699} \\
  \hline                    
  Increasing $\gamma$, 20M steps & 2570 & 13073 & 351 & 929 & 9011 \\
   & 2575 & 12873 & 361 &1031 & 10160 \\
   & 11717  & 13351 & 362 & 1099 & 9880 \\
   & 11030 & 12828 & 354 & 978 & 9263 \\
   & 2583 & 13063 & 351 & 945 & 10389 \\
\hdashline
   \multicolumn{1}{|r|}{Average} & \textbf{6095} & \textbf{13031} & \textbf{356} & \textbf{996} & \textbf{9741} \\
  \hline                    
\end{tabular}
\end{center}
\caption{Summary of the results for an increasing discount factor.} 
\label{tab:results_inc_gamma}
 \end{table}

\begin{table}[h]
\begin{center}
\begin{tabular}{  | l  | c | c | c | c | c | c |}
  \hline                       
    & Seaquest & Q-bert & Breakout & Enduro & Beam rider & Space invaders\\
  \hline                       
  Fixed $\gamma$ and $\alpha$, 50M steps & 6150 & 12958 & 401 
  & 817 & 9606 & 1976\\
  \hline                       
  Varying $\gamma$, fixed $\alpha$, 50M steps & 6016 & 13997 & 389 & 929 & 10677 & 1954 \\
  \hline                    
  Variable $\gamma$ and $\alpha$, 50M steps & 16124 & 14996 & 423 & 1129 & 12473 & 2750\\
  \hline                    
\end{tabular}
\end{center}
\caption{Summary of the results for an increasing discount factor associated with a decreasing learning rate.} 
\label{tab:results_dec_lr}
 \end{table}

\end{document}